\documentclass[10pt,twocolumn,letterpaper]{article}
\PassOptionsToPackage{table}{xcolor}

\usepackage[pagenumbers]{cvpr} 

\usepackage{threeparttable} 
\usepackage{stfloats}

\definecolor{cvprblue}{rgb}{0.21,0.49,0.74}
\usepackage[pagebackref,breaklinks,colorlinks,allcolors=cvprblue]{hyperref}

\def\paperID{25} 
\def\confName{CVPR}
\def\confYear{2025}

\title{Four Eyes Are Better Than Two: Harnessing the Collaborative Potential of Large Models via Differentiated Thinking and Complementary Ensembles}

\author{
Jun Xie$^{1,*}$, Xiongjun Guan$^{2,*,\dag}$, Yingjian Zhu$^{3}$, Zhaoran Zhao$^{1}$,\\
Xinming Wang$^{4,5}$, Hongzhu Yi$^{6}$, Feng Chen$^{1}$, Zhepeng Wang$^{1,\dag}$\\
$^1$Lenovo Research \qquad  $^2$ Tsinghua University \\
$^3$School of Artificial Intelligence, University of Chinese Academy of Sciences (UCAS) \\
$^4$Institute of Automation, Chinese Academy of Sciences(CAS) 
$^5$Zhongguancun Academy\\
$^6$University of Chinese Academy of Sciences\\
{\tt\small \{xiejun, zhaozr3, chenfeng13, wangzpb\}@lenovo.com,} \\ 
{\tt\small \{zhuyingjian24,yihongzhu23\}@mails.ucas.ac.cn,}
{\tt\small wangxinming2024@ia.ac.cn,} \\
{\tt\small gxj21@mails.tsinghua.edu.cn}
}

\begin{document}
\maketitle

\begin{NoHyper}
\def\thefootnote{}\footnotetext{* These authors contributed equally.}
\end{NoHyper}
\begin{NoHyper}
\def\thefootnote{}\footnotetext{$\dag$ Corresponding author.}
\end{NoHyper}

\begin{abstract}
In this paper, we present the \textbf{runner-up solution} for the \textbf{Ego4D EgoSchema Challenge} at CVPR 2025 (Confirmed on May 20, 2025).
Inspired by the success of large models, we evaluate and leverage leading accessible multimodal large models and adapt them to video understanding tasks via few-shot learning and model ensemble strategies.
Specifically, diversified prompt styles and process paradigms are systematically explored and evaluated to effectively guide the attention of large models, fully unleashing their powerful generalization and adaptability abilities.
Experimental results demonstrate that, with our carefully designed approach, directly utilizing an individual multimodal model already outperforms the previous state-of-the-art (SOTA) method which includes several additional processes.
Besides, an additional stage is further introduced that facilitates the cooperation and ensemble of periodic results, which achieves impressive performance improvements.
We hope this work serves as a valuable reference for the practical application of large models and inspires future research in the field.
Our Code is available at https://github.com/XiongjunGuan/EgoSchema-CVPR25.
\end{abstract}    
\section{Introduction}
\label{sec:intro}

EgoSchema \cite{mangalam2023egoschema} is a very long-form video question-answering dataset derived from Ego4D \cite{grauman2022ego4d}. 
It comprises over 5,000 manually curated multiple-choice question-answer pairs, covering more than 250 hours of real-world video data. 
For each question, the EgoSchema challenge at CVPR 2025 \cite{egovis25} requires selecting the correct answer from five given options based on a three-minute video clip.
The dataset spans a wide range of natural human activities and behaviors, characterized by long "temporal certificate" (the minimum video duration a human needs to answer the question accurately) \cite{balavzevic2024memory} and diverse complexities.
Therefore, it can serve as a diagnostic benchmark for evaluating the long-form video-language comprehension abilities of contemporary multimodal systems.

According to the training requirements, existing methods can be roughly divided into three categories: 
(1) Directly train a powerful model \cite{achiam2023gpt, papalampidi2024simple,team2024gemini,wang2024internvideo2}.
By organizing appropriate datasets and refining reward functions, these methods can flexibly align the model's expected behavior with desired trends, enabling it to better capture and comprehend specific details within certain domains.
Nevertheless, data scale and training costs are usually a significant limitation.
(2) Fine tune large models to adapt to specific tasks \cite{houlsby2019parameter,hu2022lora,cheng2024emotion}.
By focusing on training only specific modules, the overall workload is greatly reduced, while efficiently leveraging the foundational knowledge integrated during the pre-training stage.
However, there are still certain requirements for data scale, diversity, and quality.
(3) Perform prompt tuning and function assemble on accessible large models \cite{zhang2023simple,wang2023lifelongmemory,zhang2024hcqa}.
These approaches enable low-cost alignment of large models with user intent. 
Given the impressive performance of current large models, this paradigm of rapid deployment without additional training presents a highly appealing prospect.

Table \ref{tab:leaderboard} presents the performance comparison of existing works and participation methods.
From the experimental results of the first group, it can be seen that methods based on prompt tuning and model assembling \cite{wang2023lifelongmemory,zhang2024hcqa} demonstrates remarkable superiority.
Inspired by this, and considering the limited availability of labeled data (500 samples), we also adopt this training-free approach and conduct further exploration.

Unlike previous serial approaches that involve multi-stage processes \cite{wang2023lifelongmemory,zhang2024hcqa}, we conducted a systematic evaluation of prompts and chains of thought, achieving comparable and even superior performance by directly adopting end-to-end large models. 
This approach eliminates the need for cumbersome functional integration, significantly simplifying the complexity of the processing workflow. 
Furthermore, we performed an in-depth analysis of the result complementarity brought by differentiated thinking patterns and paradigms.
By leveraging this collaborative advantage through model ensemble, we further achieved a substantial improvement in accuracy.

\section{Method}
\label{sec:method}

\begin{figure*}[!t]
	\centering
	\includegraphics[width=.95\linewidth]{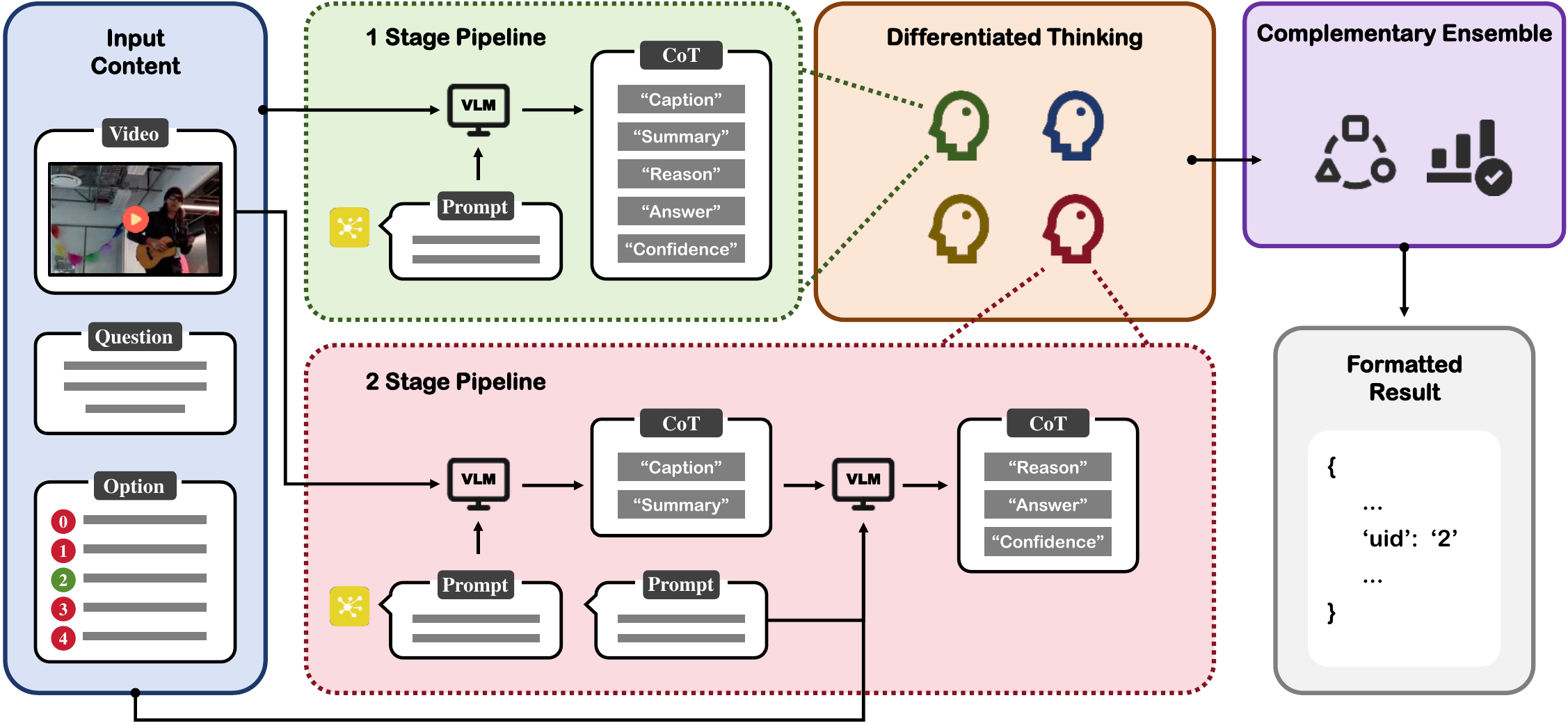}
	\vspace{1mm}
	\caption{The framework of our proposed algorithm. The specific text content is omitted in the flowchart and reported in the main text.}
	\label{fig:framework}
\end{figure*}

The overview of our proposed algorithm is shown in Figure \ref{fig:framework}.
For a given set of input content, which includes a specific video along with corresponding question and options, we construct various prompts, data streams, and Chain of Thoughts (CoT). 
These are then fed into Video-Language Models (VLMs) to generate several responses with multiple focal points.
Next, the most complementary CoT results are summarized and ensembled to determine the final answer, which is then presented in a structured and formatted manner.
The following subsections provide detailed introductions to each module, organized according to the sequence of data flow.

\subsection{Model Selection}
We first determine the candidate large VLMs that can adapt to this task.
A fundamental principle is that video description function should demonstrate strong multimodal and visual comprehension, while the response function requires robust logical reasoning and language proficiency.
According to relevant evaluations and reports, the following models have been selected and tested:
LaViLa \cite{zhao2023learning}, GPT-4o \footnote{\url{https://openai.com/index/hello-gpt-4o/}.}, Gemini 2.0 flash / 2.5 pro \footnote{\url{https://gemini.google.com/app/}.}, DeepSeek R1 / V3 \footnote{\url{https://chat.deepseek.com//}.}.

\subsection{Design of Pipeline}
Previous methods \cite{wang2023lifelongmemory,zhang2024hcqa} mostly adopted a \textbf{2-stage process}.
As shown in the red part of Figure \ref{fig:framework}, in the first stage, a VLM is used to generate a description of the video. 
In the second stage, the intermediate results and input content are input into another VLM to obtain the final result.
This approach can be considered a trade-off, as earlier VLMs lacked adequate multimodal understanding and the ability to process long-form content effectively.
Advancements in large-scale modeling technology have largely addressed these limitations.
Leveraging this progress, we develope a \textbf{1-stage process}, illustrated in the green part of Figure \ref{fig:framework}. 
Experimental results reveal that this scheme outperforms 2-stage pipeline.
One reasonable explanation is that 1-stage paradigm inherently captures all information, minimizing omissions and cumulative errors between phases. 
Moreover, the end-to-end input-output design significantly simplifies the complex multi-stage process.

\subsection{Adjustment of Prompt Words} \label{subsec:prompt}
Three styles of prompts are carefully designed and validated, including: (1) \textbf{P1}: Minimalist style text that succinctly explains requirements; 
(2) \textbf{P2}: Detailed style text that provides examples of lengthy thought processes;
(3) \textbf{P3}: Guiding style text that provides more rule guidance and moderate examples. 
The specific text content mentioned above is provided in supplementary materials (Section \ref{sup_sec:prompt}).

\subsection{Construction of CoT}
In this part, we implicitly construct the CoT process for large VLMs by specifying the items of output, thereby guiding their analysis and inference.
Specifically, five components are taken into account: captions of video clips (4s in this paper), summary of the entire video, the reason behind the decision, the answer of selected option number, and the confidence level.
In the 2-stage paradigm, the first two and the last three items serve as the outputs of the first and second stages, respectively.
On the other hand, all content needs to be provided sequentially in 1-stage paradigm.

\subsection{Model Ensemble}
By arranging and combining the aforementioned strategies, a variety of differentiated thinking outcomes can be achieved. 
Based on this foundation, highly effective and distinctive modes are selected and integrated to maximize their complementary strengths.
Let ${y_i}$ and ${\hat{y}_i}$ represent the ground truth and predicted result, respectively, the confidence level of a certain mode is calculated as
\begin{equation}
    w = \sum_i \mathbb{I} (y_i = \hat{y}_i ) \:/ \: n \;,
    \label{eq:weight}
\end{equation}
where $\mathbb{I}$ is the indicator function, $n$ is total number of samples evaluated.
The similarity between any two modes $\mathrm{A}$ and $\mathrm{A}$ is represented as
\begin{equation}
    \operatorname{sim}(\mathrm{A},\mathrm{B}) = \sum_i \mathbb{I} (\hat{y}^{\mathrm{A}}_i = \hat{y}^{\mathrm{B}}_i ) \:/ \: n \;.
    \label{eq:sim}
\end{equation}
This similarity is used to modulate the weight of each mode to avoid the influence of excessively amplifying similar opinions.
The final expression for model ensemble can be expressed as
\begin{equation}
    \operatorname{r} = \operatorname{argmax}_c\left(\sum_{k} \frac{w^k}{\sum \operatorname{sim}(\mathrm{a},\mathrm{b})} \cdot  \mathbb{I}\left(\hat{y}^k=c\right)\right)  \;,
    \label{eq:vote}
\end{equation}
where $a$, $b$ and $k$ are indexes of different modes, $c$ is the option number, $r$ is the final decision.
\section{Experiments} \label{sec:experiments}

\begin{table}[!t]
  \centering
  \begin{threeparttable} 
    \setlength{\tabcolsep}{12pt}
    \caption{Performance comparison of existing works (upper group) and the top five teams (lower group) on the public leaderboard. Our method is indicated by gray shading.}
    \label{tab:leaderboard}
    \small
    \begin{tabular}{ccc}
      \toprule
      \textbf{Method} & \textbf{Rank} & \textbf{Accuracy} \\
      \midrule
      mPLUG-Owl~\cite{ye2023mplug} $^{\dag}$ & - & 0.31 \\
      LongViViT~\cite{papalampidi2024simple} $^{\dag}$ & - & 0.33 \\
      InternVideo2~\cite{wang2024internvideo2} $^{\dag}$ & - & 0.41 \\
      LLoVi~\cite{zhang2023simple} $^{\dag}$ & - & 0.50 \\
      VideoAgent~\cite{wang2024videoagent} $^{\dag}$ & - & 0.54 \\
      ProViQ~\cite{choudhury2023zero} $^{\dag}$ & - & 0.57 \\
      Gemini 1.5 Pro~\cite{team2024gemini} $^{\dag}$ & - & 0.63 \\
      LifelongMemory~\cite{wang2023lifelongmemory} $^{\dag}$ & - & 0.68 \\
      iLearn~\cite{zhang2024hcqa} & - & 0.75 \\
      \midrule
      Noah's\_Ark\_Lab & 5 & 0.75 \\
      ccego & 4 & 0.76 \\
      iLearn2.0 & 3 & 0.77 \\
      \rowcolor{lightgray}
      L\_PCIE (PCIE) & 2 & 0.79 \\
      Reality Distortion & 1 & 0.81 \\
      \bottomrule
    \end{tabular}

    \begin{tablenotes}
      \footnotesize
      \item[$\dag$] Reported from \cite{zhang2024hcqa}.
    \end{tablenotes}
  \end{threeparttable}
\end{table}

\begin{figure}[!t]
	\centering
	\includegraphics[width=.95\linewidth]{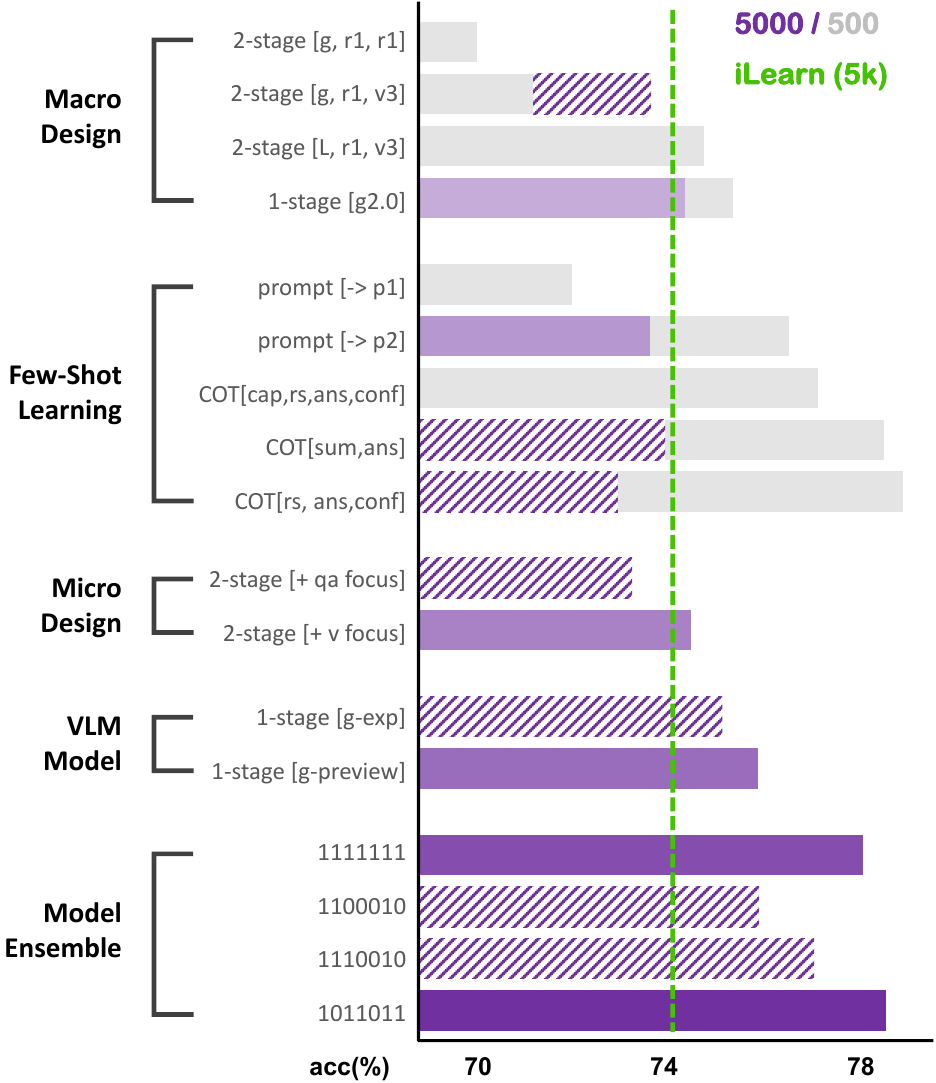}
	\caption{The comprehensive roadmap of our technological evolution. 
    The diagonal line indicates that corresponding item is not adopted in the final solution.
    The green dashed line represents the publicly available SOTA method \cite{zhang2024hcqa}.}
	\label{fig:roadmap}
\end{figure}

Table \ref{tab:leaderboard} lists the primary results of existing publicly available methods and participating teams of leaderboard.
The results demonstrate that our algorithm outperforms almost all previously proposed methods, showcasing significant improvements compared to past works proposed in related papers ($\mathbf{75\% \rightarrow 79\%}$).
At the same time, our method ranks \textbf{2nd} among non-public competition proposals, exhibiting a certain level of competitiveness.
The comprehensive roadmap of our technological evolution is presented in Figure \ref{fig:roadmap}.
Due to limited resources, our initial exploration is conducted on 500 validation samples (marked in gray) and gradually transitioned to the entire 5000 set (marked in purple).
In Section \ref{sec:ablation} of the supplementary materials, we briefly explain these ablation studies for our proposed solution, offering potential insights and inspiration for future works.

\section{Conclusion} \label{sec:conclusion}

We present our proposed runner-up solution in the Ego4D EgoSchema challenge at CVPR 2025.
Experimental results demonstrate the superiority of our approach. 
Especially, this paper details an in-depth exploration of prompts and Chain of Thought methodologies, specifically highlighting the advantages of collaboration among multiple large Vision-Language Models. 
We hope that the design philosophy and evolution of our work can provide valuable insights and inspiration for future researches.
{
    \small
    \bibliographystyle{ieeenat_fullname}
    \bibliography{main}
}


\clearpage
\setcounter{page}{1}
\maketitlesupplementary

\section{Prompt Word}
\label{sup_sec:prompt}
\begin{figure*}[b]
	\centering
	\includegraphics[width=.98\linewidth]{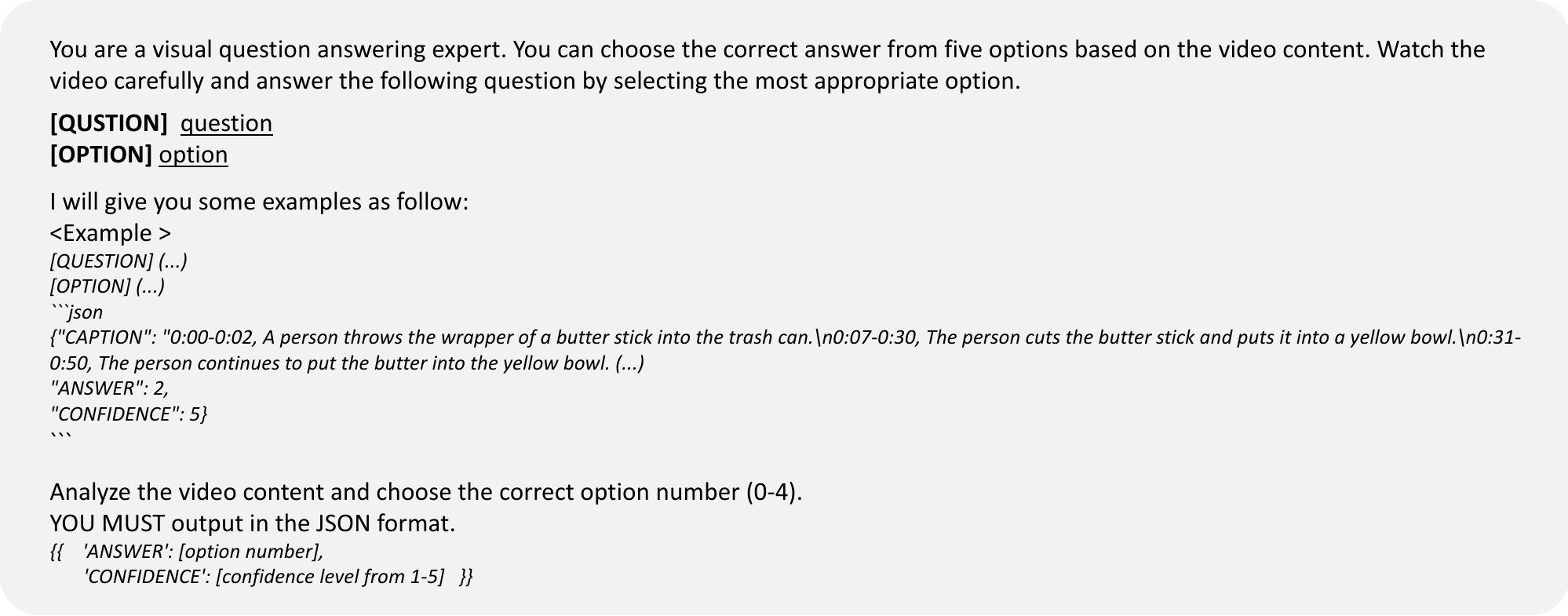}
	\caption{Example of prompt words in Style \textbf{P1}.}
	\label{fig:prompt1}
\end{figure*}

\begin{figure*}[ht]
	\centering
	\includegraphics[width=.98\linewidth]{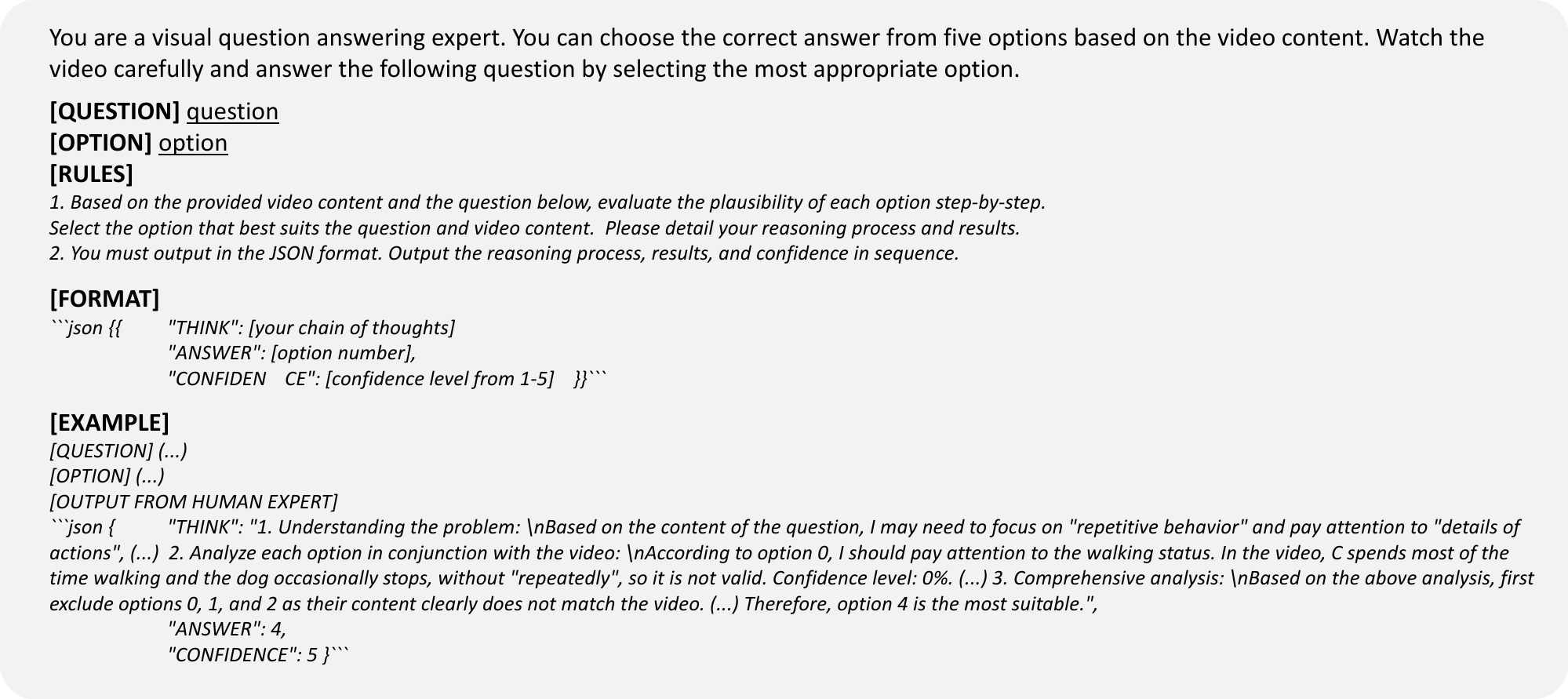}
	\caption{Example of prompt words in Style \textbf{P2}.}
	\label{fig:prompt2}
\end{figure*}

\begin{figure*}[ht]
	\centering
	\includegraphics[width=.98\linewidth]{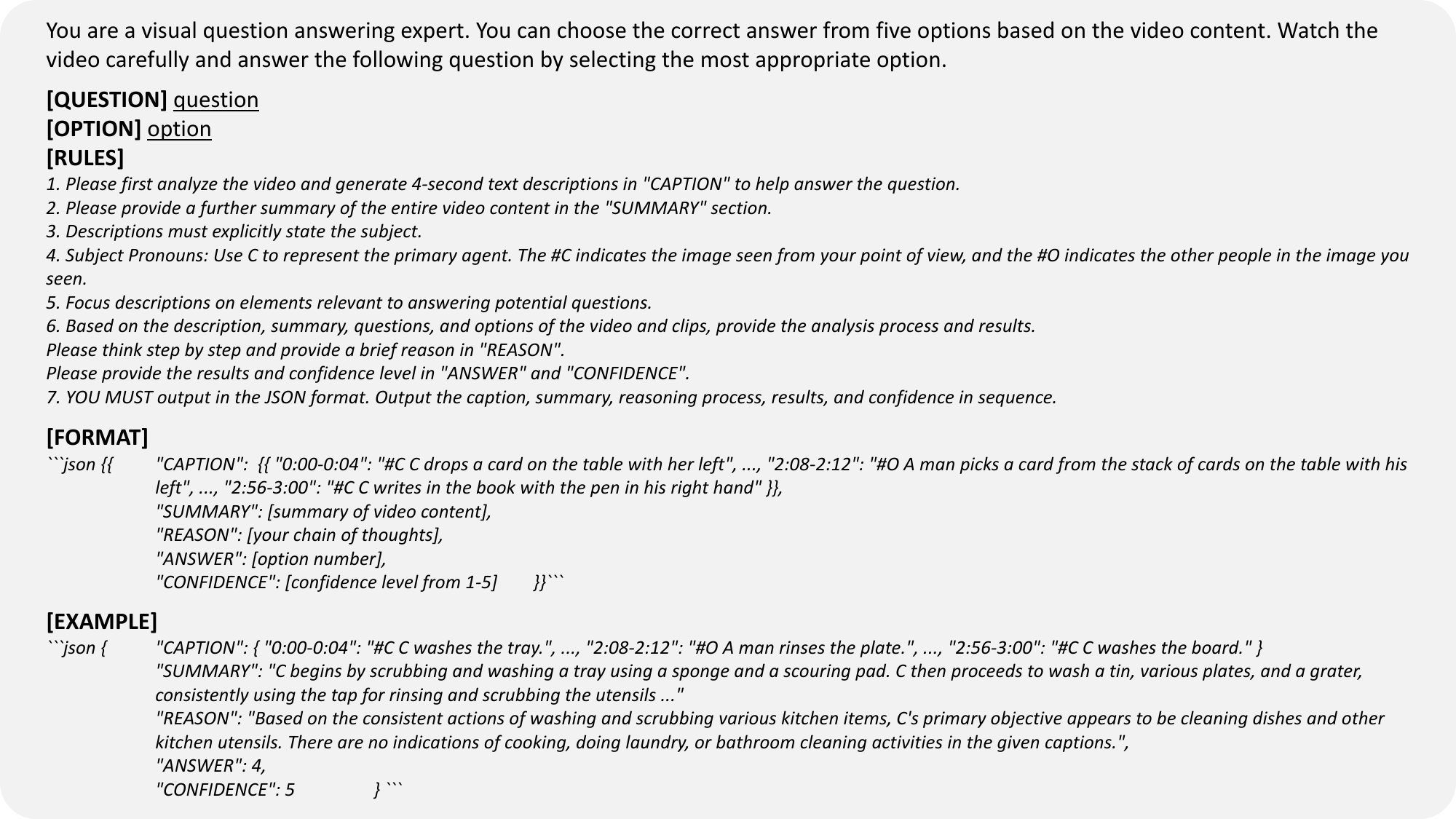}
	\caption{Example of prompt words in Style \textbf{P3}.}
	\label{fig:prompt3}
\end{figure*}

The example of prompt words in three style (introduced in section \ref{subsec:prompt}) are show in Figure \ref{fig:prompt1}, \ref{fig:prompt2} and \ref{fig:prompt3}.
Among them, the underline indicates the content that needs to be filled based on the sample.
The italics indicate the corresponding guiding content, which can promote the large model to pay attention to some task related information that may have gains.
Symbol (...) indicates that some redundant text is omitted at this location.

It should be noted that in this work, we require VLM to output JSON format for easy access to the corresponding output section.
In order to constrain the standardization of the format, the prompt explicitly provides the required output format.
Additionally, the provided example also have a positive impact on formatting to some extent.

\section{Ablation Study} \label{sec:ablation}
\subsection{Macro Design}
Our technological evolution starts with the SOTA solution, which is the 2-stage framework of iLearn \cite{zhang2024hcqa}.
The selection of VLMs and the data flow paradigm have been thoroughly evaluated.
In the top three rows, Gemini 2.0 flash (g) and LaViLa \cite{zhao2023learning} (L) are used to extract captions, respectively.
Subsequently, DeepSeek-r1 (r1) and DeepSeek-v3 (v3) are used to generate a summary from caption and perform inference in the final stage.
It can be seen that employing large models with stronger reasoning abilities during the thinking phase offers a significant improvement in accuracy, which is consistent with intuition (group line 1 \& 2).
On the other hand, LaViLa, which has been fine-tuned on this dataset, is able to generate more appropriate descriptions compared to Gemini, resulting in notable advancements (group line 2 \& 3).
This also suggests, to some extent, that in the 2-stage paradigm, the initial process of extracting descriptions from videos is susceptible to content omission or misinterpretation.
Correspondingly to this argument, our 1-stage paradigm demonstrates significant superiority (group line 4).
It should be noted that since the test results are already better than GPT-4o based method reported in \cite{zhang2024hcqa}, we did not use it in subsequent experiments due to its expensive API cost.

\subsection{Few-Shot Learning}
In this phase, the adjustment of prompt words and the construction of CoT are comprehensively evaluated.
From line 1 \& 2 in this group, it can be interpreted that excessively long text prompts in reasoning tasks might overwhelm or distract large models, leading to suboptimal performance. 
On the other hand, concise and appropriately structured prompts, combined with clear rule-based guidance, can effectively enhance the model's task-specific focus, resulting in improved outcomes.
Subsequent attempts at COT experiments demonstrate that varying modes of thinking have a significant impact on final accuracy. 
This strongly suggests that the structure and content of the output terms can heavily influence the inference performance of VLMs.
This finding underscores the importance of carefully designing the reasoning process and output format in tasks involving VLMs to achieve optimal performance.
Another interesting phenomenon is the stark difference between the trends observed in the validation set and the test set, indicating a significant data bias. 
This highlights the advantages of large models in terms of generality and adaptability, as they do not require a training process and thus avoid being affected by a small number of outlier samples.
More comparisons of CoT constructions are provided in Table \ref{tab:cot}.

\begin{table}[!t]
  \centering
  \begin{threeparttable} 
    \setlength{\tabcolsep}{9pt}
    \caption{The ablation experiment results of CoT via Gemini 2.0 flash in 500 validation dataset.
    The abbreviations in the first line, from left to right, are Caption, Summary, Reason, Answer, Confidence and Accuracy respectively.}
    \label{tab:cot}
    \small
    \begin{tabular}{c c c c c | c }
      \toprule
      \textbf{Cap} & \textbf{Sum} & \textbf{Rs} & \textbf{Ans} & \textbf{Conf} & \textbf{Acc (\%)} \\
      \midrule
      {} & {}  & {} & \checkmark & {}  & 76.0 \\
      \midrule
      \checkmark & {}  & {} & \checkmark & {}  & 75.4 \\
      {} & \checkmark  & {} & \checkmark & {}  & 78.6 \\
      {} & {}  & \checkmark & \checkmark & {}  & 77.2 \\
      {} & {}  & {} & \checkmark & \checkmark  & 77.6 \\
      \midrule
      \checkmark & \checkmark  & {} & \checkmark & {}  & 77.6 \\
      \checkmark & {}  & \checkmark & \checkmark & {}  & 77.6 \\
      {} & \checkmark  & \checkmark & \checkmark & {}  & 77.2 \\
      {} & \checkmark  & {} & \checkmark & \checkmark  & 77.6 \\
      {} & {}  & \checkmark & \checkmark & \checkmark  & 79.0 \\
      \midrule
      \checkmark & \checkmark  & \checkmark & \checkmark & {}  & 77.2 \\
      \checkmark & \checkmark  & {} & \checkmark & \checkmark  & 76.6 \\
      \checkmark & {}  & \checkmark & \checkmark & \checkmark  & 77.2 \\
      {} & \checkmark  & \checkmark & \checkmark & \checkmark  & 75.2 \\
      \midrule
      \checkmark & \checkmark  & \checkmark & \checkmark & \checkmark  & 76.6 \\
      \bottomrule
    \end{tabular}
  \end{threeparttable}
\end{table}

\subsection{Micro Design}
After thoroughly evaluating the 1-stage paradigm, we gained some valuable experience and understanding.
Based on this, we once again attempted to improve and refine the second stage paradigm.
First, we aim to utilize the VLM to generate content that may require attention based on the questions and options. 
This focus content (qa focal), along with the video and original prompt words, will then enter the first stage of the process.
On the other hand, we directly input the video into the large model and have it identify and output the parts it considers important, without providing questions or options. 
This approach aims to enhance the focus on the video itself (qa focus).
This set of results indicates that guided generation can improve targeting and efficiency, while free-form attention can enhance focus and comprehensiveness on the content itself.
Choosing the appropriate approach based on specific needs can more effectively utilize the analytical capabilities of large models.

\subsection{VLM Model}
The above experiment has conducted a basic exploration of several key aspects of large models.
In this phase, we fix the configuration of the previous optimal 1-stage solution and replace Gemini 2.0 flash with more powerful VLMs, including Gemini 2.5 exp (g-exp) and Gemini 2.5 preview (g-preview).
Naturally, the upgrade of tools brings about a significant increase in accuracy.
It is worth noting that the 2-stage paradigm has not been explored in the same manner due to the significantly higher cost of invoking APIs twice, which makes it less practical. Additionally, the coordination between stages introduces more uncontrollable interference factors, further complicating its implementation and reliability.
Therefore, we are more inclined towards a one-stage solution for practical applications.

\begin{table}[!t]
  \centering
  \begin{threeparttable} 
    \setlength{\tabcolsep}{6pt}
    \caption{Accuracy (\%) of different modes in 5000 test dataset.}
    \label{tab:acc}
    \small
    \begin{tabular}{c|ccccccc}
      \toprule
      \textbf{Mode} & 1 & 2 & 3 & 4 & 5 & 6 & 7 \\
      \midrule
      \textbf{Acc} & 75.9  & 73.7 & 75.2 & 74.0 & 73.0 & 74.4 & 73.7 \\
      \bottomrule
    \end{tabular}
  \end{threeparttable}
\end{table}

\begin{figure}[!t]
	\centering
	\includegraphics[width=.8\linewidth]{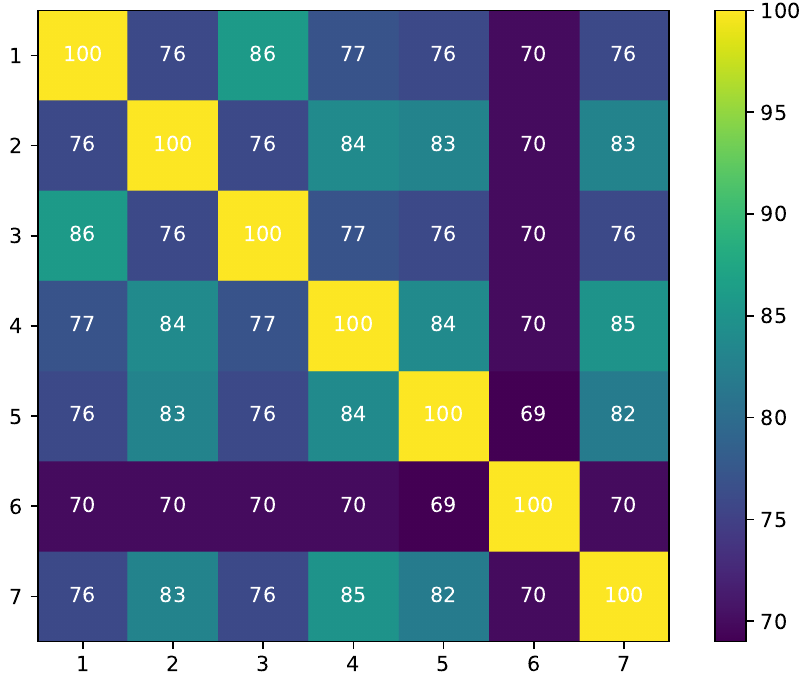}
	\caption{Similarity matrix (\%) of results from different modes in 5000 test dataset.}
	\label{fig:matrix}
\end{figure}

\subsection{Model Ensemble}
In the final phase, we select and integrate all prominent modes explored above.
A total of 7 sets of results are adopted, which have high accuracy and low similarity.
Table \ref{tab:acc} shows the accuracy (weights) of different modes calculated from Equation \ref{eq:weight}.
The visualization result of similarity matrix calculated from Equation \ref{eq:sim} is shown in Figure \ref{fig:matrix}.
We reported the main results of the assembly of these modes in the last group of Figure \ref{fig:roadmap}.
Among them, 1 represents activation, 0 represents non activation.
Several answers from activated modes are voted according to Equation \ref{eq:vote}.
Excitingly, this lightweight post-processing strategy has delivered a remarkable improvement in performance.
Furthermore, it is important to emphasize that although multiple large models were assembled, their calls were executed in parallel, and the assembly process of this model consumed virtually no time.
That is to say, our final solution is still equivalent to a 1-stage end-to-end model, maintaining both simplicity and efficiency.

\end{document}